\documentclass{article} 
\usepackage{iclr2025_conference,times}

\usepackage{amsmath,amsfonts,bm}

\def\eqref#1{equation~\ref{#1}}

\def\1{\bm{1}}

\def\rx{{\textnormal{x}}}

\def\vb{{\bm{b}}}

\def\vd{{\bm{d}}}
\def\ve{{\bm{e}}}

\def\vh{{\bm{h}}}

\def\vk{{\bm{k}}}

\def\vq{{\bm{q}}}

\def\vv{{\bm{v}}}

\def\vx{{\bm{x}}}
\def\vy{{\bm{y}}}

\def\mA{{\bm{A}}}

\def\mH{{\bm{H}}}

\def\mK{{\bm{K}}}

\def\mM{{\bm{M}}}

\def\mP{{\bm{P}}}
\def\mQ{{\bm{Q}}}

\def\mV{{\bm{V}}}
\def\mW{{\bm{W}}}
\def\mX{{\bm{X}}}
\def\mY{{\bm{Y}}}

\DeclareMathAlphabet{\mathsfit}{\encodingdefault}{\sfdefault}{m}{sl}
\SetMathAlphabet{\mathsfit}{bold}{\encodingdefault}{\sfdefault}{bx}{n}

\def\gG{{\mathcal{G}}}

\def\emA{{A}}

\def\emM{{M}}

\usepackage{hyperref}
\usepackage{url}

\title{Massive Activations in Graph Neural \newline Networks: Decoding Attention for Domain-Dependent Interpretability}

\newcommand*\samethanks[1][\value{footnote}]{\footnotemark[#1]}
\iclrfinalcopy
\author{Lorenzo Bini \thanks{These authors contributed equally.} \& Stéphane Marchand-Maillet  \\
Department of Computer Science\\
University of Geneva \\
Geneva, Carouge 1227, Switzerland \\
\texttt{\{lorenzo.bini,stephane.marchand-maillet\}@unige.ch} \\
\And
Marco Sorbi \samethanks{} \\
Research Institute for Statistics and Information Science \\
University of the Geneva \\
Geneva, Carouge 1227, Switzerland \\
\texttt{\{marco.sorbi\}@unige.ch}
}

\usepackage{graphicx}
\usepackage{subcaption}
\usepackage{cleveref}
\usepackage{amsfonts}

\begin{document}

\maketitle
 
\begin{abstract}
Graph Neural Networks (GNNs) have become increasingly popular for effectively modeling graph-structured data, and attention mechanisms have been pivotal in enabling these models to capture complex patterns. In our study, we reveal a critical yet underexplored consequence of integrating attention into edge-featured GNNs: the emergence of Massive Activations (MAs) within attention layers. By developing a novel method for detecting MAs on edge features, we show that these extreme activations are not only activation anomalies but encode domain-relevant signals. Our post‐hoc interpretability analysis demonstrates that, in molecular graphs, MAs aggregate predominantly on common bond types (e.g., single and double bonds) while sparing more informative ones (e.g., triple bonds). Furthermore, our ablation studies confirm that MAs can serve as natural attribution indicators, reallocating to less informative edges. Our study assesses various edge-featured attention-based GNN models using benchmark datasets, including ZINC, TOX21, and PROTEINS. Key contributions include (1) establishing the direct link between attention mechanisms and MAs generation in edge-featured GNNs, (2) developing a robust definition and detection method for MAs enabling reliable post-hoc interpretability.
Overall, our study reveals the complex interplay between attention mechanisms, edge-featured GNNs model, and MAs emergence, providing crucial insights for relating GNNs internals to domain knowledge.
\end{abstract}

\section{Introduction}
\label{ref:Intro}
Graph Neural Networks (GNNs) have rapidly gained traction in scientific research by effectively modeling complex graph-structured data, demonstrating remarkable success across various high-stakes applications such as bioinformatics \citep{zhang2021graph-bio}, social network analysis \citep{min2021stgsn-socialnet}, recommendation systems \citep{gao2022graph-recommender} and molecular biology \citep{cai2022fp_gnn4molecular_prediction}. In this way, understanding the internal workings of these models is crucial for ensuring their reliability and trustworthiness on such applications. Explainability in GNNs allows researchers and practitioners to identify which nodes and edges influence the model's decisions, thereby facilitating debugging, improving transparency, and building trust in the model's predictions \citep{yuan2022explainability}. Central to the recent advancements in GNNs is the integration of attention mechanisms, which enable the models to focus on the most relevant parts of the input graph, thereby enhancing their ability to capture intricate patterns and dependencies. 

Despite the substantial progress, the phenomenon of Massive Activations (MAs) \citep{MAsLLM} within attention layers has not been thoroughly explored in the context of GNNs. MAs, characterized by exceedingly large activation values, can significantly impact the stability and interpretability of neural networks. In particular, understanding and mitigating MAs in GNNs is crucial for ensuring robust and reliable model behavior, especially when dealing with complex and large-scale graphs.

However, a critical aspect of our approach lies in our deliberate choice to use edge-featured attention GNNs. These models are specifically designed to incorporate additional edge attributes,  which are typically domain-specific as chemical bond types in molecular graphs (e.g., ZINC \citep{irwin2012zinc} and TOX21 \citep{mayr2016deeptox,huang2016tox21challenge}) or spatial and interaction properties in protein graphs (e.g., PROTEINS \citep{hu2020OGB}), into their message-passing frameworks. In doing so, they attend not only to nodes but also to the rich, domain-specific information carried by edges. Conventional attention-based GNNs, such as standard Graph Attention Networks (GATs) \citep{velivckovic2017gat} and their variants that lack explicit edge-feature attention, fall outside the scope of our analysis. Our choice of models and datasets is driven by the idea that incorporating extra information at the edge-level can fundamentally alter the behavior of the attention mechanism and, consequently, the emergence of MAs.

Our central motivation is to investigate how edge-featured attention mechanisms in graph-based networks generate extreme activation values, termed MAs, which deviate from expected norms. Through empirical and statistical analyses, including the Kolmogorov–Smirnov test \citep{kolmogorov1967smirnov-test}, we demonstrate that these MAs are not only anomalies but encode domain-relevant signals (details can be found in \Cref{appedix:Further_Discussion_MAs_Detection_Procedure,KS-test}). For instance, in molecular graphs, MAs predominantly localize on common bond types (e.g., single/double bonds) rather than informative triple bonds, aligning with chemical intuition and suggesting MAs act as natural attribution indicators to highlight less informative edges. To systematically detect and characterize MAs, we develop a post-hoc interpretability framework linking edge feature integration in attention mechanisms to MA generation, alongside introducing the Explicit Bias Term (EBT) to stabilize activation distributions. Our experiments comprehensively evaluate GNN architectures, GraphTransformer \citep{Bresson-GT}, GraphiT \citep{GraphiT}, and SAN \citep{SAN}, across diverse tasks (graph regression, multi-label classification) to validate the consistency of MAs. By establishing MA identification criteria and conducting ablation studies, we underscore the role of edge features in shaping these activations, thereby offering actionable insights for model interpretation and stabilization. While our current analysis provides a deep characterization of MAs, we remain committed to further exploring additional datasets and configurations in future work.

In summary, our contributions are twofold \footnote{Code is public available on our GitHub \href{https://github.com/msorbi/gnn-ma}{page}.}:
\begin{itemize}
    \item We provide the first systematic study on MAs in edge-featured attention-based GNNs, highlighting their impact on model interpretability.
    \item We propose a robust detection methodology for MAs, accompanied by detailed experimental protocols and ablation studies to enable reliable post‐hoc interpretability of model attention outputs.
\end{itemize} Through this work, we aim to shed light on a critical yet understudied aspect of attention-based GNNs, offering valuable insights for the development of more interpretable graph-based models.

\section{Related Works}
\label{sec:Related-Works}
GNNs have emerged as powerful tools for analyzing graph-structured data, with applications in healthcare \citep{paul2024systematic-gnns4healthcare}, molecular property prediction \citep{wieder2020compact-gnn4molecular-property}, and computational biology discovery \citep{bini2024flowcyt}. The evolution of GNNs has seen significant advancements, particularly with the integration of attention mechanisms inspired by transformers in natural language processing \citep{vaswani2017attention}. GATs \citep{velivckovic2017gat} pioneered the use of self-attention in GNNs, enabling nodes to dynamically weigh their neighbors, thereby enhancing the model's ability to capture complex graph relationships. Subsequent innovations, such as GraphiT \citep{GraphiT} and the Structure-Aware Network (SAN) \citep{SAN}, further generalized transformer architectures for graphs and incorporated structural properties, improving performance across tasks. 

Recent studies on Large Language Models (LLMs) and Vision Transformers (ViTs) have identified the presence of extreme activation values (MAs) in their attention layers \citep{xiao2023efficient, MAsLLM, darcet2023vision, dosovitskiy2020image}, prompting investigations into their implications for model behavior, interpretability, and robustness. While similar phenomena have been observed in ViTs, the study of MAs in GNNs remains underexplored, representing a critical gap in understanding these models. 

Broader research on neural network interpretability, such as feature visualization \citep{Olah2017distill} and network dissection \citep{bau2017network-dissection}, offers potential methodologies for analyzing MAs in GNNs. Additionally, insights from attention flow \citep{abnar2020attentionflow} and attention head importance \citep{michel2019sixteen-heads} in transformers suggest that not all attention heads contribute equally, raising questions about similar patterns in graph transformers and their relation to MAs. These findings highlight the need for further research into MAs in GNNs to uncover their role, impact, and potential vulnerabilities.
The study of internal representations in deep learning models has been a topic of significant interest in the machine learning community. Works such as \citet{bau2020understanding} have explored the interpretability of neural networks by analyzing activation patterns and their relationships to input features and model decisions. However, the specific phenomenon of MAs in GNNs has remained largely unexplored until now, representing a crucial gap in our understanding of these models and their relationships to the domain of the data they process.

\section{Establishing the Reference}
\begin{figure*}[t]
    \centering
    \begin{subfigure}{0.49\linewidth}
        \includegraphics[width=\linewidth]{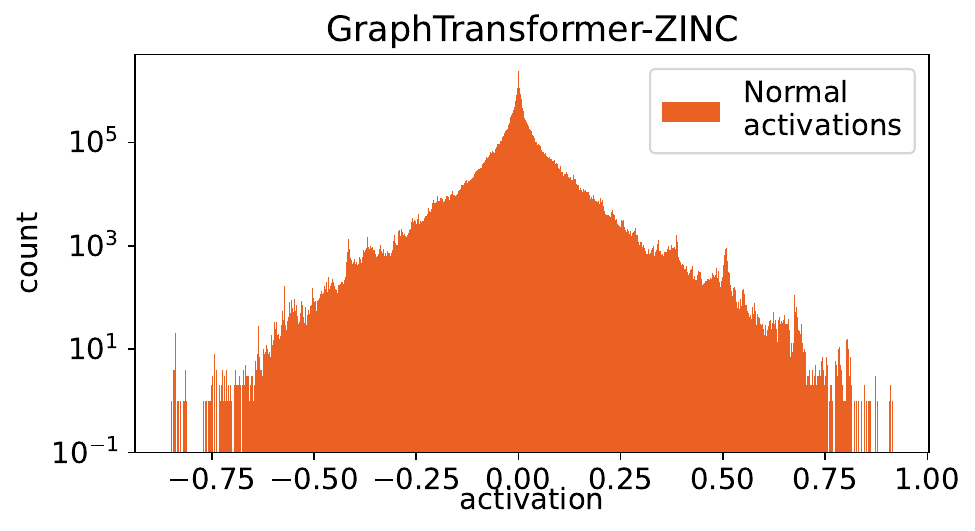}
        \caption{Base model}
        \label{fig:zero_distribution}
    \end{subfigure}
    \hfill
    \begin{subfigure}{0.49\linewidth}
        \includegraphics[width=\linewidth]{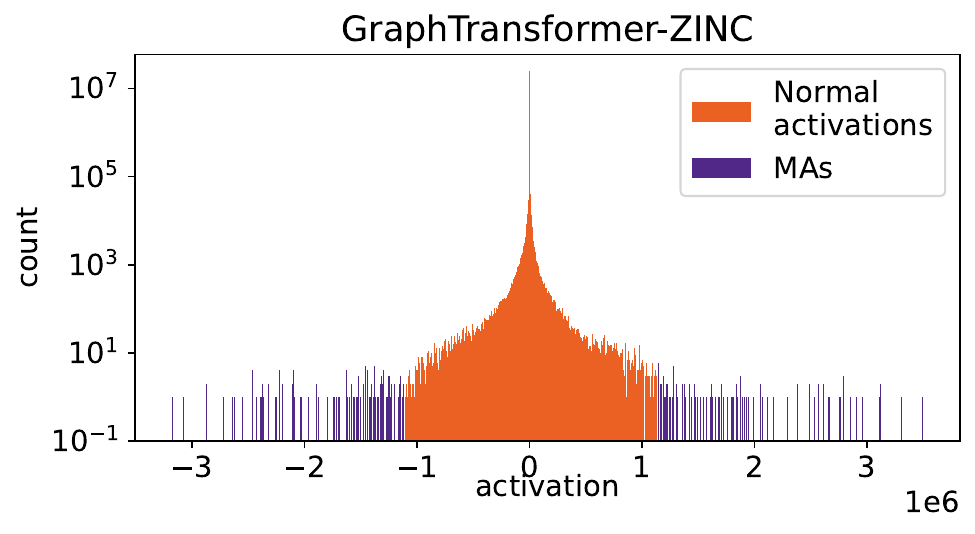}
        \caption{Trained model}
        \label{fig:MA_raw_distrs}
    \end{subfigure}
    \caption{Comparison of activation distributions between base and trained model. The trained model presents two long tails representing MAs values. Notice the $y$-axis is log-scaled, making the base model activations distribution clustered around zero.}
    \label{fig:zero_distribution_all}
\end{figure*}

In this section, we detail our approach to analyzing activation distributions in attention-based GNNs, emphasizing a dual perspective: an untrained baseline analysis and a-posteriori observation of a distribution shift in trained models, mapped as outlier activations. We begin by stabilizing a controlled baseline to establish an interpretability reference. This baseline serves as a litmus test for detecting and quantifying deviations and outliers in trained models, as explained in \Cref{sec:Terminology-MA,sec:Methodology-Observations}. In their initialized state, attention values follow a symmetric, near-zero distribution (\Cref{fig:zero_distribution}), a consequence of standard weight initialization schemes. This initial behavior embodies our expectations for the model's internal dynamics before any task-specific training occurs.
We start by considering the untrained (base) model, where network parameters are initialized via Xavier initialization \citep{glorot2010xavier}. 
To form a meaningful baseline, we normalize the activation values within each layer. Specifically, for each edge activation, we compute the ratio:
\begin{equation}
\label{eq:ratio}
\text{ratio}(\text{activation}) = \frac{|\text{activation}|}{\text{median}(|\text{edge activations}|)}.
\end{equation} This normalization, dividing by the layer’s edge median, accounts for scale variations across layers and models. To facilitate a meaningful analysis, we apply a logarithmic transformation to the activations ratio (\Cref{eq:ratio}). This transformation exposes the intrinsic shape of the activation distribution, making subtle differences more discernible. As illustrated in \Cref{fig:zero_distribution}, the resulting base distribution is highly peaked, with the majority of values clustered around zero, yet exhibits a long tail for higher values. This sharp peak serves as a robust baseline, reflecting the model's inherent activation scale before any training-induced changes occur. In this state, the model has not yet learned task-specific features and the activations predominantly reflect the properties of the random initialization. 

Our choice is to model the log-transformed base distribution as a Gamma distribution. This decision is motivated by both theoretical and empirical observations. The Gamma distribution, a flexible two-parameter family, is well-suited to capture the skewed, unimodal behavior that arises from the logarithmic transformation of \Cref{eq:ratio}. In the untrained (base) model these transformed activation values are well-captured by the Gamma distribution. Empirically, as shown in \Cref{fig:gammaa}, our analysis demonstrates that the negative log-transformed activation ratios from the base model align closely with the Gamma approximation. This is validated by a very low Kolmogorov-Smirnov (KS) statistic (approximately $0.020$), confirming that the Gamma distribution accurately reflects the statistical properties of the base activations. Thus, both theoretical suitability and strong empirical fit justify the use of the Gamma to model the base activation distribution.

Before delving into the modeling of the distribution shift, it is important to bridge our analysis from the established baseline to the observation of training-induced changes. In the untrained (base) model, as described above, the baseline serves as our reference point for understanding the activation behavior before any task-specific learning occurs. However, as the model is trained, its internal dynamics evolves significantly, as later shown in \Cref{sec:Terminology-MA}. By comparing the base and trained models in \Cref{fig:zero_distribution_all}, we observe that activation profile exhibits anomalous concentrations on the left and right tails. As depicted in \Cref{fig:distrs}, while the Gamma distribution accurately approximates the base activations, it fails to capture the extreme values, i.e. MAs appearing after training (which correspond to left-hand values due to the application of log-transformation). This two-part framework, beginning with an initial baseline and progressing to a post-hoc investigation, ensures our analysis not only captures the behavior of the base model but also offers explainable insights into the modifications induced by training. In \Cref{sec:Terminology-MA} we introduce the appropriate definitions and terminology for MAs. Then, throughout \Cref{sec:Methodology-Observations} we proceed with the investigation of the training-corrupted distribution and the consequences of the MAs' emergence. 

\section{Terminology of Massive Activations in GNNs}
\label{sec:Terminology-MA}
Building upon the work on MAs in LLMs \citep{MAsLLM}, we extend this investigation to edge-featured attention-based GNNs, focusing specifically on graph transformer architectures. Our study encompasses various models, including GraphTransformer (GT) \citep{Bresson-GT}, GraphiT \citep{GraphiT}, and Structure-Aware Network (SAN) \citep{SAN}, applied to diverse task datasets such as ZINC, TOX21, and OGBN-PROTEINS (see \Cref{Model-Architecure,Datset-Composition} for details on models configurations and datasets composition). This comprehensive approach allows us to examine the generality of MAs across different attention-based GNN architectures.

\subsection{Characterization of Massive Activations}
MAs in GNNs refer to specific activation values that exhibit unusually high magnitudes compared to the typical activations within a layer. These activations are defined by the following criterion, where an activation value is intended to be its absolute value.


\textbf{Relative Threshold}: In the paper by \citet{MAsLLM}, MAs were defined as at least 1,000 times larger than the median activation value within the layer. This relative threshold criterion helped differentiate MAs from regular high activations that might occur due to normal variations in the data or model parameters. The formal definition was represented as $\text{MAs} = \{a \mid a > 1000 \times \text{median}(\mathbf{A})\}$, where $\mathbf{A}$ represents the set of activation values in a given layer.
However, in contrast to previous studies that employed a fixed relative threshold to detect LLMs MAs, our work is intended to characterize their nature within an a-posteriori explainable framework. This investigation ensures a comparative analysis of the GNNs attention activations, where the untrained model serves as a reference to identify emerging outliers.

\subsubsection{Detection Procedure}
For both base and trained models, we detected MAs following a systematic procedure:

\textbf{Normalization}: We normalized the activation values within each layer, dividing them by the edge median on the layer, to account for variations in scale between different layers and models. This normalization step ensures a consistent basis for comparison. Since attention is computed between pairs of adjacent nodes only, in contrast to LLMs where it is computed among each pair of tokens, the model tends to spread MAs among the edges to make them ``available'' to the whole graph. Indeed, our prior analysis indicates that MAs are a common phenomenon across different models and datasets, that they are not confined to specific layers but are distributed throughout the model architecture, and that MAs are an inherent characteristic of the attention-based mechanism in graph transformers and related architectures, not strictly dependent on the choice of the dataset (see \Cref{appedix:Further_Discussion_MAs_Detection_Procedure} for further details, in particular \Cref{fig:MA-edges}). 
\begin{figure*}[t]
    \centering
    \includegraphics[width=0.9\linewidth]{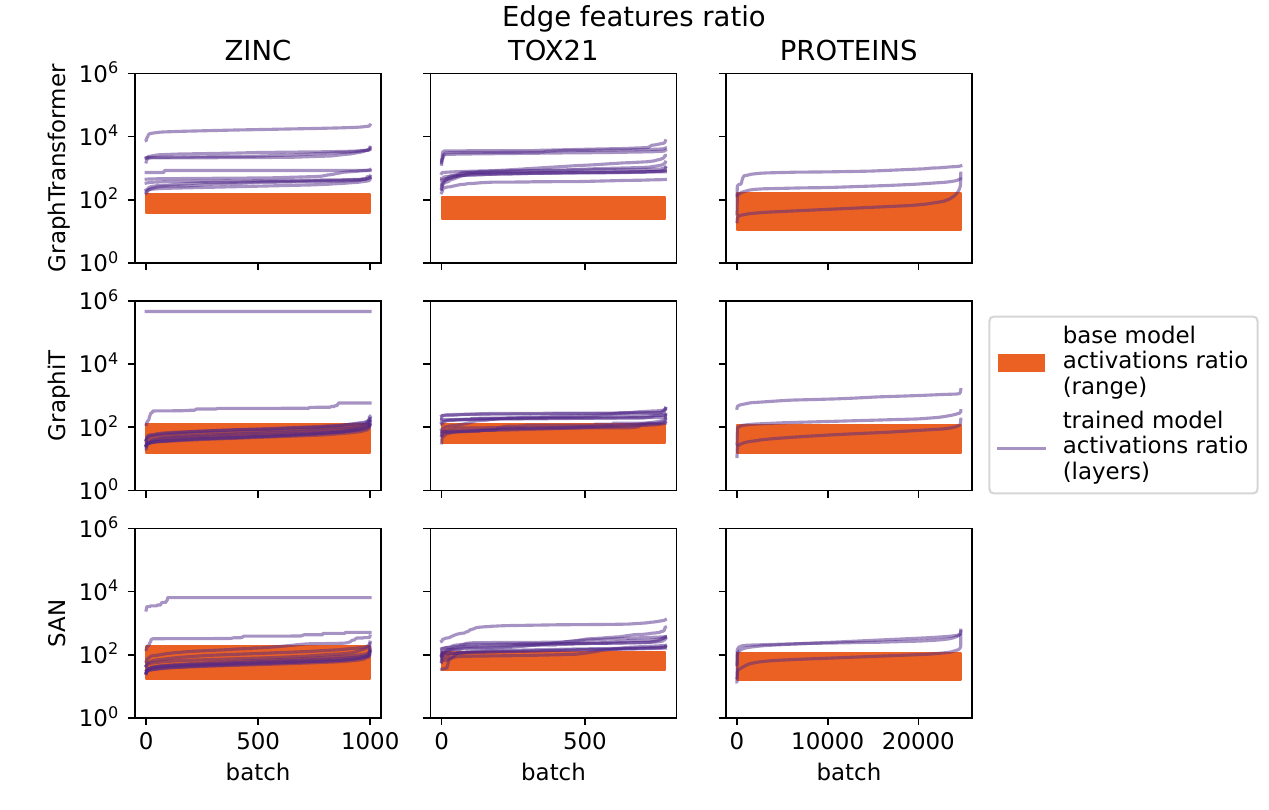}
    
    \caption{Comparison of MAs on trained against base models. Represented ratios have been sorted increasingly for each layer independently. Orange box represents the range of normal ratios obtained in the base model, while ratios exceeding the base come from MAs.}
    \label{fig:MA-nobias}
\end{figure*}

\textbf{Batch Analysis}: We analyzed the activations on a batch-by-batch basis, minimizing the batch size, to have suitable isolation between the MAs and to ensure that the detection of MAs is not influenced by outliers in specific samples. 
For each activation we computed its ratio as in \Cref{eq:ratio}, and those exceeding the threshold were flagged as massive. We then considered the maximum ratio of each batch to detect those containing MAs. We performed this analysis across multiple layers to identify patterns and layers that are more prone to exhibiting MAs. This aggregation helps in understanding the hierarchical nature of MAs within the model.

\Cref{fig:MA-nobias} reports the analysis results. The batch ratios significantly increase in the trained transformers, concerning base ones, often even overcoming the threshold of 1000 defined by previous works \citep{MAsLLM}, showing the presence of MAs in graph transformers.

\section{Methodology and Observation}
\label{sec:Methodology-Observations}
Focusing on edge features, first, we analyzed the ratio defined in \Cref{eq:ratio}, taking the maximum for every batch, across each model layers, and visually compared the outcomes to value ranges obtained using the same model in a base state (with parameters randomly initialized, without training) to verify the appearance of MAs. The graphical comparison, reported in \Cref{fig:MA-nobias}, shows ratios over the base range in most of the trained models, representing MAs. 

\begin{figure*}[t]
    \centering
    \begin{subfigure}{0.43\linewidth}
        \includegraphics[width=\linewidth]{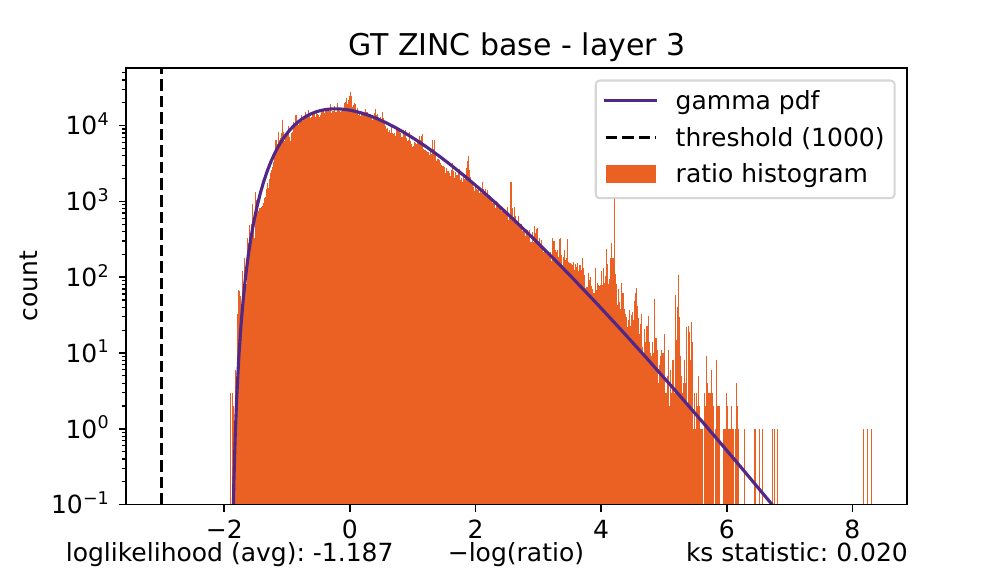}
        \caption{Good approximation (Base)}
        \label{fig:gammaa}
    \end{subfigure}
    \begin{subfigure}{0.43\linewidth}
        \includegraphics[width=\linewidth]{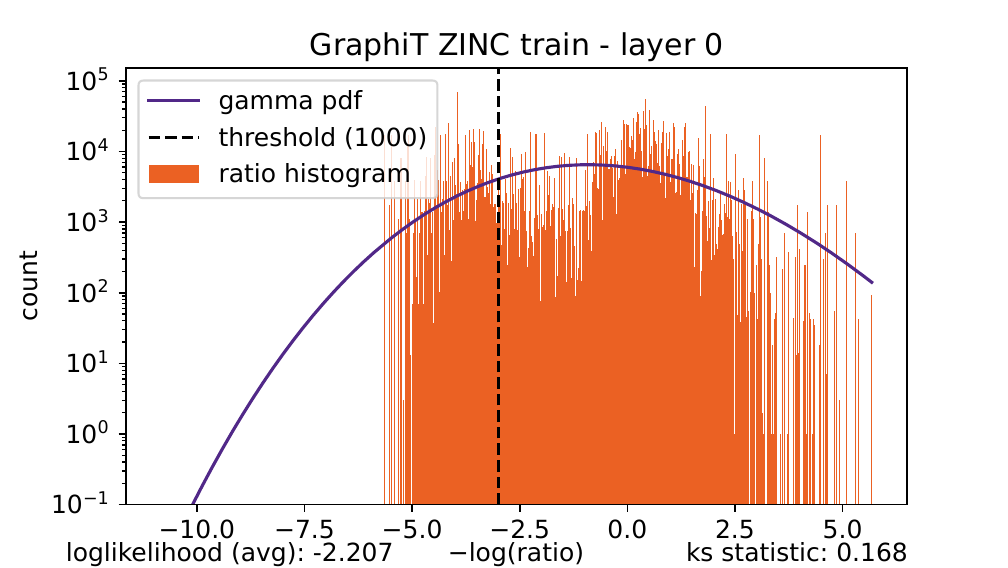}
        \caption{Bad approximation (MAs)}
        \label{fig:gammab}
    \end{subfigure}
    \hfill
    \begin{subfigure}{0.43\linewidth}
        \includegraphics[width=\linewidth]{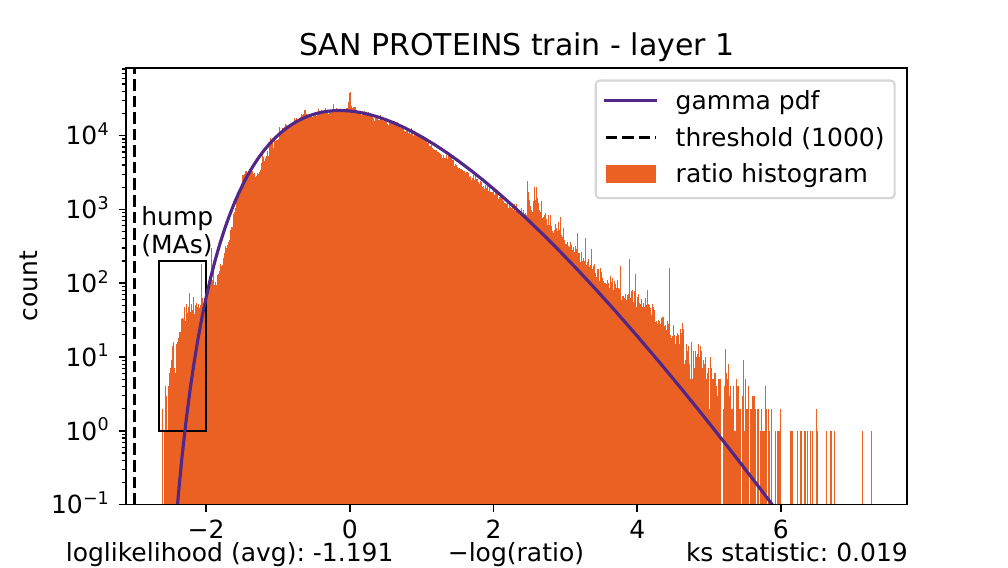}
        \caption{Good approximation (MAs)}
        \label{fig:gammac}
    \end{subfigure}
    \begin{subfigure}{0.43\linewidth}
        \includegraphics[width=\linewidth]{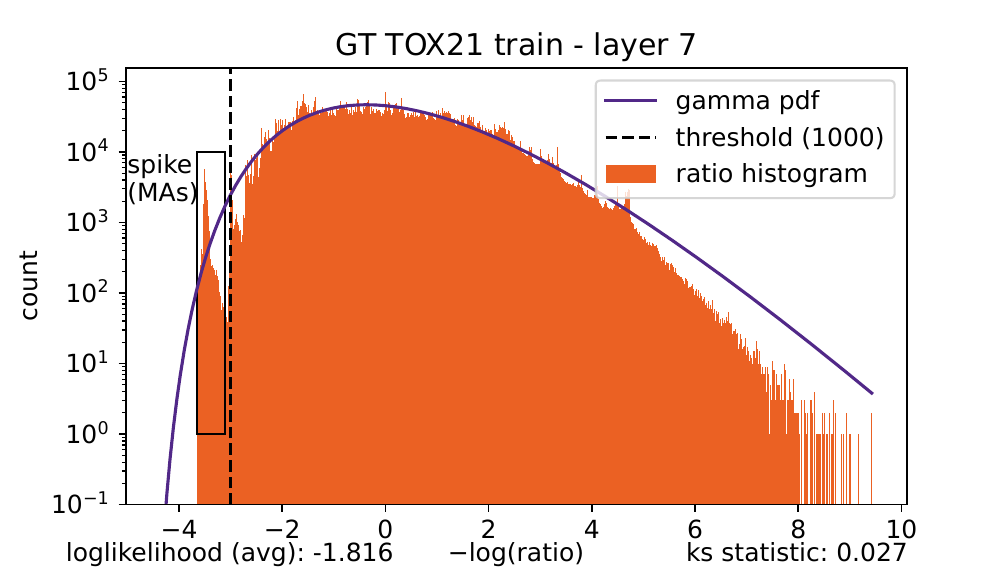}
        \caption{Bad approximation (MAs)}
        \label{fig:gammad}
    \end{subfigure}
    
    \caption{Activation distributions for base and trained (with MAs) models. In \Cref{fig:gammad} we clearly distinguish a spike on the left of the distribution, corresponding to a ratio of 1000 (-$\log(\text{ratio}) = -3$), which identifies the separation between the basic and massive regimes. The approximation pdf is rescaled to match the histogram scale.}
    \label{fig:distrs}
\end{figure*}

To better characterize MAs, we studied their distribution employing the Kolmogorov-Smirnov statistic \citep{kolmogorov1967smirnov-test}, as discussed in \Cref{KS-test}. We found that a gamma distribution well approximates the negative logarithm of the activations' magnitudes, as well as their ratios. \Cref{fig:gammaa} shows this approximation for a base model layer.
We point out that, according to the existing definition, items on the left of the $-3$ are MAs.
We compared the distributions of the $\log$-values between the base and trained models, as illustrated in \Cref{fig:distrs}, which highlights a significant shift in the trained model's distribution, confirming the emergence of MAs during training. This shift indicates that the threshold around $-\log(\text{ratio}) = -3$ (e.g., a ratio of 1000 or higher) effectively captures these significant activations, though it sometimes appears slightly shifted to the right, as shown in \Cref{fig:gammac}. 

When MAs appear, two phenomena are observed: either a large number of extreme activation values are added to the left-hand side of the distribution, preventing a good approximation (\Cref{fig:gammab}), or a few values appear as spikes, humps, or out-of-distribution values, which may or may not deteriorate the approximation (\Cref{fig:gammad,fig:gammac}). For instance, \Cref{fig:gammaa} represents the base model with untrained weights, where the gamma approximation fits the sample histogram well, evidenced by a low KS statistic of $0.020$. In contrast, \Cref{fig:gammab} shows the trained model's distribution with a significant shift due to a large hump on the left side, representing extreme activation ratios (MAs), resulting in a poor gamma approximation with a KS statistic of $0.168$. Similarly, \Cref{fig:gammad} displays a clear spike at $-\log(\text{ratio}) = -3$ (a ratio of 1000) in the trained model's distribution, indicating the distinction between basic and massive activation regimes and a poor gamma fit with a KS statistic of $0.027$. Finally, \Cref{fig:gammac} shows the trained model's distribution with a noticeable hump on the left side, indicating MAs. Although the gamma approximation fits better here (KS statistic of $0.019$), the presence of MAs is still evident, confirming their addition to the left-hand side of the distribution.

Inspired by recent advancements in addressing bias instability in LLMs \citep{MAsLLM}, we introduced an EBT into our graph transformer models. This bias term is discovered to counteract the emergence of MAs by stabilizing the activation magnitudes during the attention computation. The EBT is computed as follows:
\begin{align}
    \vb_e &= \bm{Q} \vk \ve' \label{eq:bias_e} \\
    \vb_v &= \text{softmax}(\bm{A}_e) \vv' ,  \label{eq:bias_v}
\end{align} where $\vk, \ve, \vv \in \mathbb{R}^{d}$ are the key, edge, and node bias terms (one per each attention head), $\bm{A}_e$ is the edge attention output, and $d$ the corresponding hidden dimension. $\vb_e$ and $\vb_v$ represent the edge and node bias terms and are added to the edge and node attention outputs, respectively. By incorporating EBT into the edge and node attention computations, and adding bias in the linear projections of the attention inputs, we regulated the distribution of activation values, thus mitigating the occurrence of MAs. Further details on the MA detection procedure and EBT's impact are available in \Cref{appedix:Further_Discussion_MAs_Detection_Procedure}.

In the next section, we delve into the interpretability of edge-related MAs, demonstrating how their emergence provides insights into the model’s attention allocation. By analyzing MAs in relation to domain-specific edge features, we reveal their role as natural attribution indicators. This investigation highlights how MAs can be leveraged to understand and refine graph transformer models, improving their interpretability and facilitating their use in scientific discovery.

\section{Interpretability of Edge-Related Massive Activation}
\label{sec:interpretability_heatmap}
\begin{figure*}[t]
    \centering
    \includegraphics[width=0.763\linewidth]{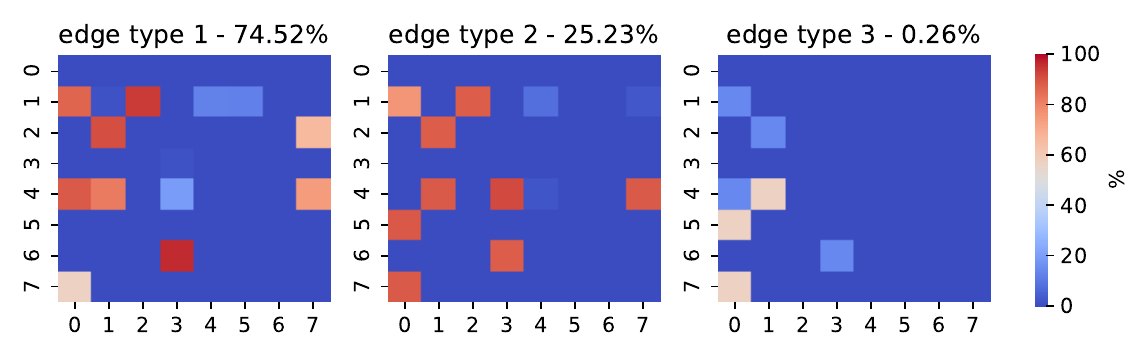}
    \caption{Heatmaps showing MAs concentration across the three edge types in the ZINC dataset. Each heatmap visualizes the percentage of edges with MAs per attention head and hidden feature dimension. Notably, MAs predominantly aggregate on edge types 1 and 2, while being absent on type 3, indicating the model's tendency to allocate MAs to more frequent edges. Type $3$ consists of $0.26\%$ of edges in the dataset.}
    \label{fig:heatmap}
\end{figure*}

The emergence of MAs raises critical questions about \textit{why} and \textit{where} these outliers occur in graph structures. In the context of molecule graphs, we analyze MAs through the lens of edge types, a human-interpretable graph feature, and quantify their role in driving model behavior. We employ edge type-wise activation heatmap to localize MAs within the graph topology. In the ZINC dataset, edge types represent different types of chemical bonds between atoms in a molecule, specifically edge type 1 corresponds to a single bond (e.g., C$-$H), edge type 2 represents a double bond (e.g., C$=$O), and edge type 3 indicates a triple bond (e.g., C$\equiv$C ). Triple bonds are less common but highly significant in certain chemical contexts. For each edge type, we explain the model's attention output through a heatmap (\Cref{fig:heatmap,fig:heatmap_ablation}), where we visualize MAs per attention head and hidden feature dimension. Specifically, each cell in the heatmap represents the percentage of edges having one MA in that position. For example, \Cref{fig:heatmap_ablation} heatmap with edge type $5$, shows that at position $(7,0)$ $100\%$ of edges have one MA each on that location. \Cref{fig:heatmap} reveals a distinguished pattern: MAs are aggregated on edge types $1$ and $2$, and not present on type $3$. This observation provides several critical insights into the model’s internal behavior:
\begin{itemize}
    \item The aggregation of MAs on edge types $1$ and $2$ indicates that the model has a particular regard for most rare edges type.
    \item Under normal conditions, without the influence of MAs, the activation values on each edge would depend on the ``token'' contextual information. However, the presence of MAs introduces extreme values that overwrite these domain-dependent signals.
    \item In accordance with Shannon information \citep{shannon1948mathematical}, a higher frequency of occurrence is generally associated with lower per-instance information content, as the information becomes more diffusely distributed. Broadly, given an event $x$ with probability $P$, the information content is defined as $I(X):= -\log_2[\text{Pr(x)}] = -\log_2(P)$. In this way, type $3$ edges (less frequent) are most informative ones.
    \item The model appears to have learned to identify less informative edges and exploit them to allocate MAs, thereby leaving unmodified original domain information on critical edges.
\end{itemize} These insights suggest MAs can serve as edge importance indicator to retrieve domain-relevant information. For instance, in self-supervised/contrastive learning scenarios, 
rather than solely relying on hand-crafted augmentations (which may be suboptimal for certain tasks) one could design augmentation strategies leveraging MAs as indicators. Leveraging these indicators can be beneficial for downstream tasks, where identifying critical edges, those that significantly influence the model's performance, is essential for creating meaningful augmentations. Measures like link entropy \cite{brandes2001faster,dehmer2011history} and graph cuts \cite{shin2022graphcut} can be employed to assess the importance of edges \citep{qian2017quantifying,li2024comprehensive_community_detection}, guided by MAs as indicators for deploying augmentation strategies to improve learning efficiency. 

It is important to clarify that the significance of an edge is not uniquely determined by its type; rather, it depends on contextual information and graph structure as well \citep{borgatti2005centrality,vzalik2023density}. For our current analysis, however, we have focused on investigating the relationship between edge type and MAs presence. 

\subsection{Ablation Studies on the Interpretability of Edge-Related MAs}
\label{sec:ablation_interpretability}
\begin{figure*}[t]
    \centering
    \includegraphics[width=\linewidth]{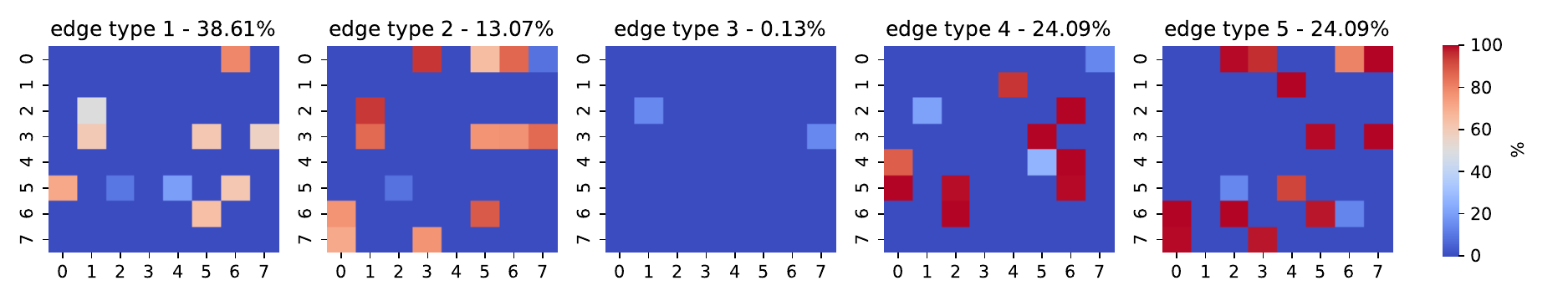}
    \caption{Ablation studies show the reallocation of MAs on the chemically meaningless edges (types 4 and 5), designed to carry low intrinsic information. This supports the hypothesis that MAs can serve as indicators of edge importance.}
    \label{fig:heatmap_ablation}
\end{figure*}

To further investigate our use of MAs as indicators of less informative edges, we conducted an ablation study designed to decouple chemical informativeness from edge frequency. In our experiment, for each molecule in the dataset we introduced a global dummy node that connects to all other atoms. This connection is established through two new types of edges: type $4$ for incoming connections to the dummy node and type $5$ for outgoing connections. As a result, while the most frequent edge type (i.e., single chemical bond) remains type $1$, the newly introduced edges (types $4$ and $5$) are intentionally meaningless from a chemical standpoint and thus represent edges with very low intrinsic information content.
This controlled setup allows us to clearly observe that the network, once retrained, reallocates MAs towards dummy edges (types $4$ and $5$) designed to be less informative, as shown in \Cref{fig:heatmap_ablation}. This reallocation confirms our hypothesis that MAs serve as markers for edges carrying lower domain-specific information content. Such findings suggest that MAs could be exploited as indicators of edge importance to guide downstream tasks.
\section{Conclusion and Future Work}
In this work, we have presented the first study of MAs in edge-featured attention-based GNNs. Our novel methodology for detecting and analyzing MAs, supported by ablation studies, has demonstrated that these extreme activations are not model artifacts but can be linked with edge importance. By establishing a robust framework for post-hoc interpretability, we have shown that MAs provide valuable insights into how attention mechanisms allocate importance across edges, revealing, for example, that common bond types in molecular graphs tend to accumulate these activations while more informative bonds remain relatively unaltered. This work thus not only deepens our understanding of the internal mechanisms of edge-featured attention GNNs but also sets the stage for their application in extracting actionable scientific insights.
Furthermore, our investigation highlights the role of EBT in stabilizing activation distributions.

Looking forward, our future work will expand this interpretability framework across a broader range of architectures and datasets. We aim to further explore how MAs patterns can be systematically exploited to improve model transparency and guide the design of data-adaptive strategies for downstream tasks such as link prediction, drug design, and self-supervised learning. By investigating how measures like edge entropy relate to MAs distribution, we plan to refine augmentation and feature re-weighting techniques that enhance both model performance and interpretability.

In summary, our study provides a key step towards developing more transparent and interpretable graph-based models. By addressing the challenges posed by MAs and leveraging them as natural attribution indicators, we aim to bridge the gap between complex neural network internals and domain-specific scientific discovery.


\section*{Acknowledgments}
The Swiss National Science Foundation partially funds this work under grants number 207509 "Structural Intrinsic Dimensionality", and 215733 "Une édition sémantique et multilingue en ligne des registres du Conseil de Genève (1545-1550)".

\bibliography{iclr2025_conference}
\bibliographystyle{iclr2025_conference}

\newpage
\appendix
\section{Dataset Composition}
\label{Datset-Composition}
This section provides additional details on the used datasets throughout the experiments.

The ZINC dataset \citep{irwin2012zinc} is a benchmark collection for evaluating GNNs in molecular chemistry, where molecules are represented as graphs with atoms as nodes and chemical bonds as edges. Contents include:
\begin{itemize}
    \item Graphs: The dataset includes over $250,000$ molecular graphs. Each molecule is represented by a graph with nodes (atoms) and edges (bonds), incorporating various bond types (e.g., single, double, triple).
    \item Node Features: Atoms are described by features that capture their chemical properties, such as atom types, hybridization states, and other atomic attributes.
    \item Edge Features: Bonds between atoms are characterized by features representing bond types and additional chemical information.
    \item Task: The primary task is \textbf{graph regression}, where the goal is to predict continuous values associated with each molecule. This often involves predicting molecular properties such as solubility or biological activity.
\end{itemize} ZINC \cite{irwin2012zinc} is useful for evaluating GNNs' performance in learning molecular representations and predicting continuous chemical properties, providing insights into the model’s ability to generalize across diverse chemical compounds.

The TOX21 dataset \citep{mayr2016deeptox, huang2016tox21challenge} is designed for toxicity prediction and focuses on classifying chemical compounds based on their potential toxicity. It is part of the Toxicology Data Challenge and features molecular graphs with associated toxicity labels. Contents include:
\begin{itemize}
    \item Graphs: The dataset consists of molecular graphs where nodes represent atoms and edges represent chemical bonds. It includes thousands of molecules with toxicity annotations, and it consists of $7,831$ graphs with each graph representing a molecular structure with associated toxicity labels.
    \item Node Features: Atoms are encoded with features representing their types, hybridization states, and other chemical properties.
    \item Edge Features: Bonds are detailed with features indicating bond types and additional chemical attributes.
    \item Task: The main task is \textbf{multi-label graph classification}, where each molecule is classified into multiple toxicity categories. This allows for the prediction of various toxicity endpoints simultaneously.
\end{itemize} TOX21 \citep{mayr2016deeptox,huang2016tox21challenge} is valuable for assessing GNN models in predicting toxicity from molecular structures, which is crucial for drug discovery and safety evaluation, providing a benchmark for multi-label classification tasks.

The OGBN-PROTEINS dataset, part of the Open Graph Benchmark (OGB) \citep{hu2020OGB}, focuses on protein function prediction. It contains one large graph representing protein structures, with nodes corresponding to amino acids and edges to their interactions. Contents include:
\begin{itemize}
    \item One Large Graph: OGBN-PROTEINS contains $54,879$ nodes and $89,724$ edges. These nodes represent amino acids in protein structures, and edges represent interactions or bonds between these amino acids. It includes various protein structures used for functional prediction.
    \item Node Features: Amino acids are described by features capturing biochemical properties, such as amino acid type, secondary structure, and other relevant attributes.
    \item Edge Features: Edges denote interactions between amino acids and include features reflecting the nature of these interactions or spatial relationships.
    \item Task: The task is \textbf{multi-label node classification}, where the goal is to predict multiple functional categories for each amino acid node in the protein graph. This involves classifying nodes into various functional classes based on their role in the protein's functionality.
\end{itemize} OGBN-PROTEINS \citep{hu2020OGB} is suitable for evaluating GNNs on biological data, specifically in predicting protein functions based on structural information. It provides insights into how well models can handle multi-label node classification tasks in a complex biological context.

\section{Further Discussion on MAs Detection Procedure}
\label{appedix:Further_Discussion_MAs_Detection_Procedure}
\begin{figure*}[t]
    \centering
    \includegraphics[width=0.9\linewidth]{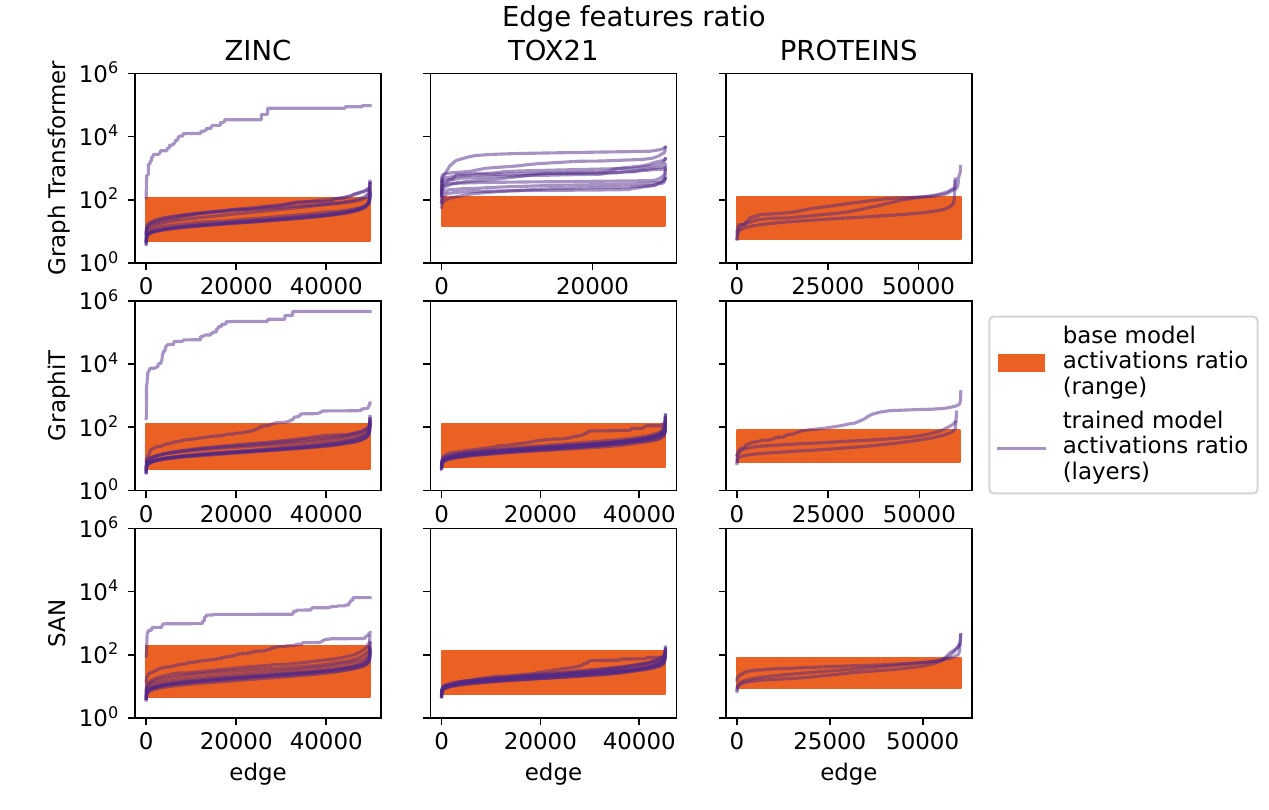}
    \caption{Comparison of MAs for trained vs base models, along all the edges. Activation values have been normalized within each layer by the layer's edge median. Represented ratios have been sorted increasingly for each layer independently.}
    \label{fig:MA-edges}
\end{figure*}

The analysis presented in \Cref{sec:Methodology-Observations} highlights key insights into the emergence and distribution of MAs in edge-featured attention-based GNNs. As illustrated in \Cref{fig:MA-edges,fig:MA-nobias}, distinct patterns emerge across datasets and model architectures, revealing the interplay between attention mechanisms, dataset characteristics, and learned biases. Below, we summarize the main findings drawn from our evaluation.
\begin{figure*}[t]
    \centering
    \includegraphics[width=0.9\linewidth]{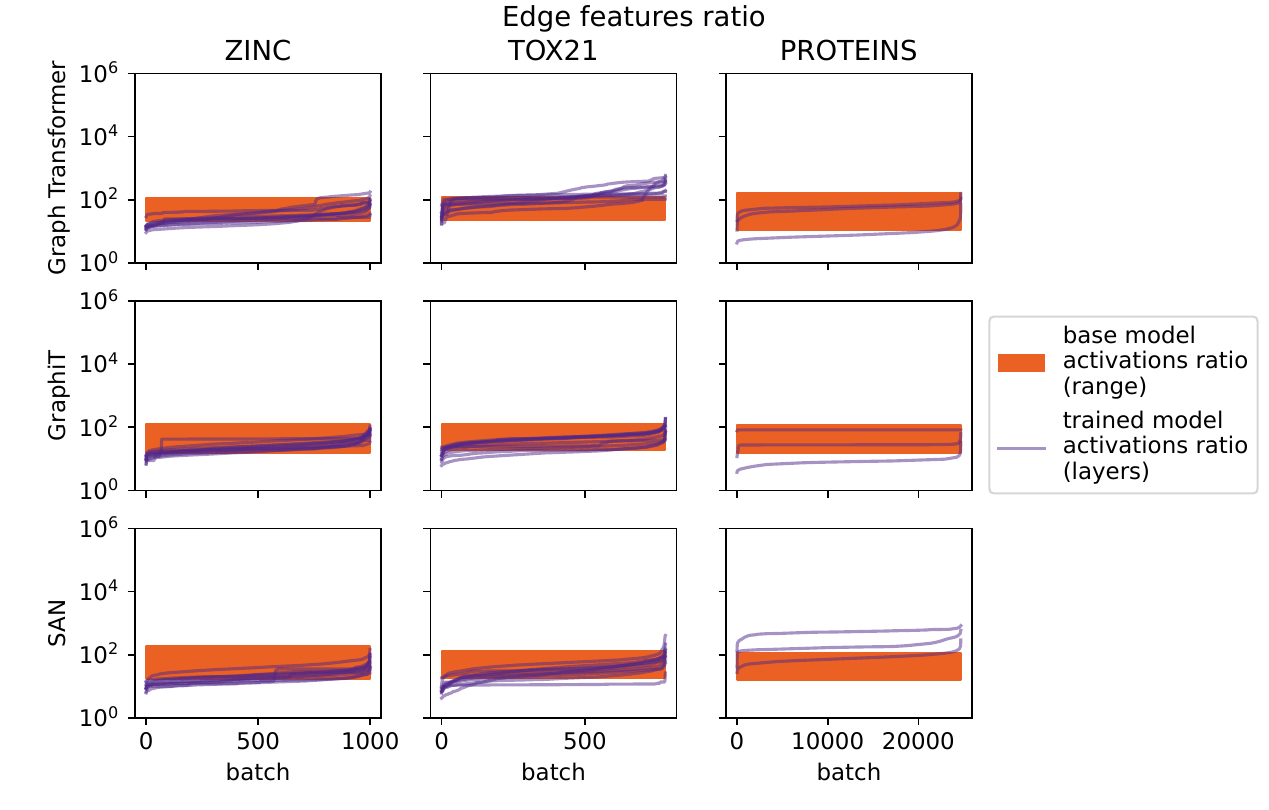}
    \caption{Comparison of MAs on trained against base models, with the use of Explicit Bias Term. Represented ratios have been sorted increasingly for each layer independently.}
    \label{fig:MA-bias}
\end{figure*} 
\begin{enumerate}
    \item \textbf{Dataset Influence}:
    \begin{itemize}
        \item The ZINC and OGBN-PROTEINS datasets consistently show higher activation values across all models compared to TOX21, suggesting that the nature of these datasets significantly influences the emergence of MAs. Even though many MAs are emerging form GT on TOX21.
    \end{itemize}
    \item \textbf{Model Architecture}:
    \begin{itemize}
        \item Different GNN models exhibit varying levels of MAs. For instance, GraphTransformer and GraphiT tend to show more pronounced MAs than SAN, indicating that model architecture plays a crucial role.
    \end{itemize}
    \item \textbf{Impact of Attention Bias}:
    \begin{itemize}
        \item Previous works suspect that MAs have the function of learned bias, showing that they disappear introducing bias at the attention layer. This holds for LLMs and ViTs, and for our GNNs as well, as shown in \Cref{fig:MA-nobias} where the presence of MAs is affected by the introduction of the Explicit Bias Term on the attention. \Cref{fig:MA-bias} and text below suggest that MAs are intrinsic to the models' functioning, being anti-correlated with the learned bias.
    \end{itemize}
\end{enumerate}
The consistent observation of MAs in edge features, across various GNN models and datasets, points to a fundamental characteristic of how these models process relational information.
\Cref{tab:test_loss} shows that EBT does not systematically influence the test loss equally across different models and datasets. We have considered the test loss metric to keep the approach general, making it extendable to different downstream tasks. This ensures that the proposed method can be applied broadly across various applications of graph transformers. 

Although the test loss remains relatively unchanged with the introduction of EBT, its presence helps in mitigating the occurrence of MAs, as evidenced by the reduction in extreme activation values observed in earlier figures. By analyzing these results, it becomes evident that while EBT does not drastically alter the test performance, it plays a crucial role in controlling activation anomalies, thereby contributing to the robustness and reliability of graph transformer models. 

As illustrated in \Cref{fig:MA-bias}, the introduction of EBT leads to a substantial reduction in both the frequency and magnitude of MAs, aligning activation ratios more closely with those seen in the base models. This stabilization effect is consistently observed across all datasets, ZINC, TOX21, and OGBN-PROTEINS, demonstrating that EBT effectively regulates activation distributions, bringing them closer to the expected reference behavior of untrained models.
This consistency underscores the general applicability of EBT in various contexts and downstream tasks. Moreover, \Cref{fig:MA-bias} shows that EBT mitigates MAs across different layers of the models. This is crucial as it indicates that EBT's effect is not limited to specific parts of the network but is extended throughout the entire architecture. For example, GraphTransformer on ZINC without EBT shows MAs frequently exceed $10^4$, while when EBT has been applied these ratios are significantly reduced, aligning more closely with the base model's range.
\begin{table}[t]
\caption{Comparison of test loss with and w/o bias for the different models and datasets. In bold the worst performances.}
\label{tab:test_loss}
\begin{center}
\begin{tabular}{llll}
\textbf{Dataset} & \textbf{Model} & \textbf{Test loss} & \textbf{Test loss (EBT)}
\\ \hline \\
ZINC 
& GraphTransformer & 0.26 & \textbf{0.29} \\
& GraphiT & 0.13 & \textbf{0.31} \\
& SAN & 0.18 & \textbf{0.27} \\
\\
TOX21 
& GraphTransformer & 0.25 & \textbf{0.29} \\
& GraphiT & \textbf{0.38} & 0.32 \\
& SAN & \textbf{0.38} & 0.31 \\
\\
OGBN-PROTEINS 
& GraphTransformer & \textbf{0.13} & 0.12 \\
& GraphiT & 0.14 & \textbf{0.16} \\
& SAN & 0.13 & 0.13 \\
\end{tabular}
\end{center}
\end{table}

\section{Kolmogorov-Smirnov Test}
\label{KS-test}
This section provides additional details on the Kolmogorv-Smirnov (KS) test \citep{kolmogorov1967smirnov-test} used to analyze the distribution of activations. The KS test is a non-parametric test that compares the cumulative distribution functions of two samples. It is used to compare a sample with a reference probability distribution (one-sample KS test) or to compare two samples (two-sample KS test) with each other. We primarily used the one-sample KS test to assess the goodness of fit between our observed activation distributions and a theoretical gamma distribution.

In our study, we utilized the KS statistic to compare the distribution of activation values before and after training (i.e. base against trained model), identifying MAs. Xavier initialization was chosen  due to its well-established ability to maintain stable activation distributions throughout deep networks, reducing the risk of vanishing or exploding gradients. As shown in \Cref{fig:zero_distribution_all}, the distribution observed in the untrained model is the closest approximation to a Delta function among all cases, with activations concentrated around their expected mean (zero). This serves as a crucial reference for assessing how training and the emergence of MAs alter the model’s internal behavior. Once training begins, learned weights and attention mechanisms introduce deviations from this distribution. 

\subsection{One-Sample Kolmogorov-Smirnov Test}
The one-sample KS test can typically be formulated as follows:
\subsubsection{Null Hypothesis}
The null hypothesis for the one-sample KS test is:

$H_0$: The sample data follows the specified distribution (in our case, a gamma distribution).

\subsubsection{Test Statistic}
The KS statistic $D_n$ is defined as the supremum of the absolute difference between the empirical cumulative distribution function (ECDF) $F_n(\rx)$ of the sample and the cumulative distribution function (CDF) $F(\rx)$ of the reference distribution:

\begin{equation}
    D_n = \sup_x |F_n(x) - F(x)|
\end{equation}
where $\sup_x$ denotes the supremum of the set of distances.

\subsubsection{Empirical Cumulative Distribution Function}
For a given sample $x_1, x_2, ..., x_n$, the ECDF is defined as:

\begin{equation}
    F_n(x) = \frac{1}{n} \sum_{i=1}^n \1_{x_i \leq x}
\end{equation}
where $\1_{x_i \leq x}$ is the indicator function, equal to $1$ if $x_i \leq x$ and $0$ otherwise.

\subsubsection{Critical Values and p-value}
The distribution of the KS test statistic under the null hypothesis can be calculated, which allows us to obtain critical values and p-values. The null hypothesis is rejected if the test statistic $D_n$ is greater than the critical value at a chosen significance level $\alpha$, or equivalently if the p-value is less than $\alpha$.

\subsection{Application to MAs Detection}
\label{sec:Application2MAs}
In our experiments, we used the KS statistic to assess whether the distribution of activation ratios in our GNNs follows a gamma distribution. The process is as follows:

\begin{enumerate}
    \item We computed the activation ratios for each layer of our models, as defined in \Cref{eq:ratio} of the main paper.
    \item We took the negative logarithm of these ratios to transform the distribution.
    \item We fit a gamma distribution to this transformed data using maximum likelihood estimation.
    \item We performed a one-sample KS test to compare our sample data to the fitted gamma distribution.
\end{enumerate}

The KS test statistic provides a measure of the discrepancy between the observed distribution of activation ratios and the theoretical gamma distribution. A lower KS statistic indicates a better fit, suggesting that the activation ratios more closely follow the expected distribution.

\subsection{Interpretation in the Context of MAs}

Following the described procedure in Section \ref{sec:Application2MAs}, we employed the KS statistic as quantitative/statistical measure to detect the presence of MAs:

\begin{itemize}
    \item For untrained (base) models, we typically observed low KS statistics, indicating that the activation ratios closely follow a gamma distribution.
    \item For trained models exhibiting MAs, we often saw higher KS statistics. This indicates a departure from the gamma distribution, which we interpret as evidence of MAs.
    \item The magnitude of the KS statistic provided a quantitative measure of how significantly the presence of MAs distorts the expected distribution of activation ratios.
\end{itemize} Moreover, we complemented our KS statistic results with visual inspections of the distributions and other analyses as described in the main paper.




\section{Model Architecture}
\label{Model-Architecure}
This section provides additional details on the models' architecture used throughout all the experiments, namely GT \citep{Bresson-GT}, GraphiT \citep{GraphiT} and SAN \citep{SAN}. These graph-transformer architectures integrate the principles of both GNNs and transformers, leveraging the strengths of attention mechanisms to capture intricate relationships within graph-structured data. Graph transformers extend the transformer structure, typically used for sequence data, to graphs, operating by embedding nodes and edges into higher-dimensional spaces and then applying multi-head self-attention mechanisms to capture dependencies between nodes.

Mathematically, let $\gG = (V, E)$ be a graph where $V = \{v_1, ..., v_n\}$ is the set of nodes and $E \subseteq V \times V$ is the set of edges. Each node $v_i$ is associated with a feature vector $\vx_i \in \mathbb{R}^d$, and each edge $(v_i, v_j)$ may have an edge feature $\ve_{ij} \in \mathbb{R}^k$. Therefore, graph transformer models are designed as follows.

\subsubsection*{Input Embedding}
The initial node features $\mX = [\vx_1, ..., \vx_n]^T \in \mathbb{R}^{n \times d}$ are typically projected to a higher-dimensional space:
\begin{equation}
    \mH^{(0)} = \mX\mW_{in} + \vb_{in}
\end{equation}
where $\mW_{in} \in \mathbb{R}^{d \times d'}$ is a learnable weight matrix and $\vb_{in} \in \mathbb{R}^{\vd'}$ is a bias vector.

\subsubsection*{Positional Encoding}
To capture structural information, positional encodings $\mP \in \mathbb{R}^{n \times d'}$ are often added:
\begin{equation}
    \mH^{(0)} = \mH^{(0)} + \mP
\end{equation}

\subsubsection*{Multi-Head Attention Layer}
The core of a graph transformer is the multi-head attention mechanism. For each attention head $i$ (out of $h$ heads) there are also:

\begin{enumerate}
    \item Query, Key, and Value Projections:
    \begin{align}
    \mQ_i &= \mH^{(l)}\mW_i^Q \\
    \mK_i &= \mH^{(l)}\mW_i^K \\
    \mV_i &= \mH^{(l)}\mW_i^V
\end{align} where $\mW_i^Q, \mW_i^K, \mW_i^V \in \mathbb{R}^{d' \times d_k}$ are learnable weight matrices, and $d_k = d' / h$.
    \item Attentions Scores (node features only): \begin{equation}
    \mA_i = \text{softmax}\left(\frac{\mQ_i \mK_i^T}{\sqrt{d_k}} + \mM\right) ,
\end{equation}
where $\mM \in \mathbb{R}^{n \times n}$ is a mask matrix to enforce the graph structure:
\begin{equation}
    \emM_{i,j} = \begin{cases} 
        0 & \text{if } (v_i, v_j) \in E \text{ or } i = j \\
        -\infty & \text{otherwise} .
    \end{cases}
\end{equation}
    \item Output of each head:
\begin{equation}
    \textbf{head}_i = \mA_i \mV_i .
\end{equation}
    \item Concatenation and Projection: \begin{equation}
    \mH' = \text{Concat}(\textbf{head}_1, ..., \textbf{head}_h)\mW^O ,
\end{equation}
where $\mW^O \in \mathbb{R}^{d' \times d'}$ is a learnable weight matrix.
\end{enumerate}

\subsubsection*{Feed-Forward Network (FFN)}
Each attention layer is typically followed by a position-wise feed-forward network:
\begin{equation}
    \text{FFN}(\vx) = \max(0, \vx\mW_1 + \vb_1)\mW_2 + \vb_2
\end{equation}
where $\mW_1 \in \mathbb{R}^{d' \times d_{ff}}$, $\mW_2 \in \mathbb{R}^{d_{ff} \times d'}$, $\vb_1 \in \mathbb{R}^{d_{ff}}$, and $\vb_2 \in \mathbb{R}^{d'}$ are learnable parameters.

\subsubsection*{Layer Normalization and Residual Connections}
Each sub-layer (attention and FFN) employs a residual connection followed by layer normalization:
\begin{equation}
    \mH^{(l+1)} = \text{LayerNorm}(\mH^{(l)} + \text{Sublayer}(\mH^{(l)}))
\end{equation}
where Sublayer is either the multi-head attention or the FFN.

\subsubsection*{Edge Feature Integration}
GraphTransformer, GraphiT and SAN incorporate edge features:
\begin{enumerate}
    \item In attention computation: \begin{equation}
    \emA_{i,j} = \text{softmax}\left(\frac{\vq_i^T \vk_j + f(\ve_{ij})}{\sqrt{d_k}}\right)
\end{equation}
where $f$ is a learnable function (e.g., a small neural network) that projects edge features.
    \item In value computation: \begin{equation}
    \vv_{ij} = \mV_i + g(\ve_{ij})
\end{equation}
where $g$ is another learnable function.
\end{enumerate}

\subsubsection*{Global Node}
Some architectures introduce a global node $v_g$ connected to all other nodes to capture graph-level information:
\begin{equation}
    \vh_g^{(l+1)} = \text{Attention}(\vh_g^{(l)}, \mH^{(l)})
\end{equation}

\subsubsection*{Output Layer}
The final layer depends on the task:
\begin{itemize}
    \item For node classification: $\vy_{node} = \text{softmax}(\mH^{(L)}_{node}\mW_{out} + \vb_{out})$
    \item For graph classification: $\mY_{graph} = \text{MLP}(\text{Pool}(\mH^{(L)}))$
\end{itemize}
where Pool is a pooling operation (e.g., mean, sum, or attention-based pooling) to switch from single node to graph embedding level.

\subsubsection*{Training}
The model is typically trained end-to-end using backpropagation to minimize a task-specific loss function, such as cross-entropy for classification or mean squared error for regression.

\end{document}